\documentclass[format=acmsmall, screen=true, review=false, nonacm=true]{acmart}

\AtBeginDocument{%
  \providecommand\BibTeX{{%
    \normalfont B\kern-0.5em{\scshape i\kern-0.25em b}\kern-0.8em\TeX}}}

\usepackage{subfig}
\usepackage{tikz}
\usetikzlibrary{positioning, calc, shapes.geometric, shapes.symbols, arrows.meta, fit, backgrounds, intersections}
\usepackage{pgfplots}
\pgfplotsset{compat=1.16}
\usepackage{pgfplotstable}

\usepackage{multirow}

\usepackage{xstring}

\usepackage{lmodern}
\begin{document}

\title{Keep Calm and Relax}
\subtitle{HMI for Autonomous Vehicles}


\author{Tima~M.~Yekta}
\authornote{Both authors contributed equally to this research.}
\email{fmohammadiye@uos.de}
\affiliation{%
  \institution{Osnabrück University}
  \department{Institute of Cognitive Science}
  \city{Osnabrück}
  \country{Germany}
}

\author{Julius Schöning}
\authornotemark[1]
\email{j.schoening@hs-osnabrueck.de}
\affiliation{%
  \institution{Osnabrück University of Applied Sciences}
  \department{Faculty of Engineering and Computer Science}
  \city{Osnabrück}
  \country{Germany}
}


\begin{abstract}
The growing popularity of self-driving, so-called autonomous vehicles has increased the need for human-machine interfaces~(HMI) and user interaction~(UI) to enhance passenger trust and comfort. While fallback drivers significantly influence the perceived trustfulness of self-driving vehicles, fallback drivers are an expensive solution that may not even improve vehicle safety in emergency situations. Based on a comprehensive literature review, this work delves into the potential of HMI and UI in enhancing trustfulness and emotion regulation in driverless vehicles. By analyzing the impact of various HMI and UI on passenger emotions, innovative and cost-effective concepts for improving human-vehicle interaction are conceptualized. To enable a trustful, highly comfortable, and safe ride, this work concludes by discussing whether HMI and UI are suitable for calming passengers down in emergencies, leading to smarter mobility for all.
\end{abstract}








\maketitle

\section{Introduction}
Trust and autonomous vehicles are two words that increased the demand for user interaction~(UI) and human-machine interface~(HMI) design. Trust is one of the essential factors in autonomous and automated technology~\cite{Anderson2021,Shahrdar2018} and will become more urgent in driverless vehicles in the upcoming decade. Human drivers have become superfluous within driverless, automated, and self-driving vehicles. Thus, in the cases of fully and highly automated driving, trust can no longer be provided by the presence of a driver or a fallback driver. The driving system itself must provide trust. Especially where these vehicles are required to operate in challenging environments and unexpected scenarios, the innovative and cost-effective HMI and UI must enhance passenger trust by, e.g., emotion regulation~\cite{Generosi2022,Braun2021}. Active emotion regulation in inevitable fear and stress-provoking situations increased the trust, comfort, and pleasure in self-driving vehicles. By effectively addressing passengers' emotional needs and expectations, automotive UI and HMI enable trustworthy autonomous vehicles in sustainable transportation systems.

This paper explores the potential of HMI and UI in achieving trust and emotion regulation for self-driving vehicles. Through a comprehensive literature review, this work examines the impact of various HMI and UI concepts on passenger emotions and proposes innovative solutions for improving human-vehicle interaction. Finally, the paper discusses whether HMI and UI are suitable for calming passengers down in emergencies and challenging environments to enable smart and trustful mobility.

\section{User Interaction Clusters in Partial and Full Automated Vehicles} 
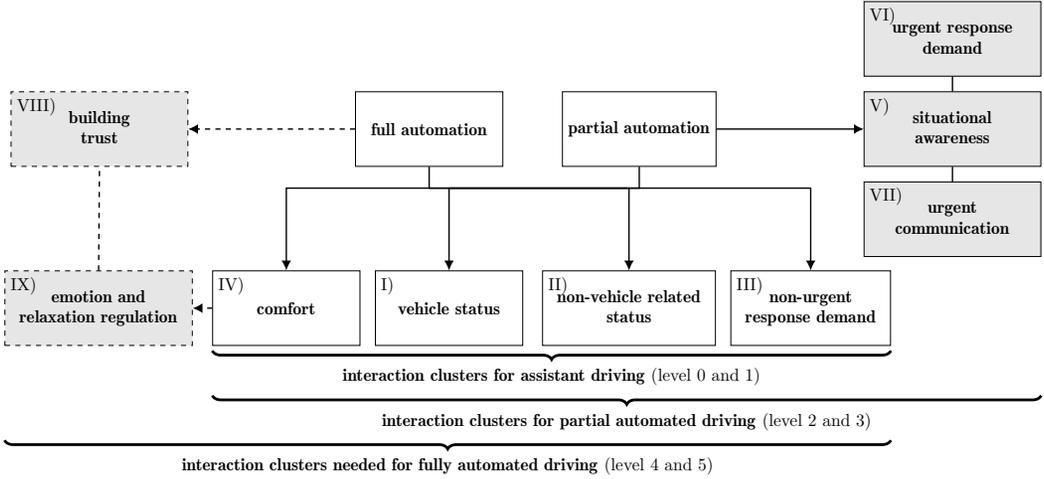
\begin{figure*}[tb]

\resizebox{\linewidth}{!}{
\begin{tikzpicture}
  \node(fullAuto)[ draw, rectangle, minimum width = 3.0cm, minimum height = 1.5cm] {\textbf{full automation}};
  \node(partAuto)[right = 1.2cm of fullAuto, draw, rectangle, minimum width = 3.0cm, minimum height = 1.5cm] {\textbf{partial automation}};

  \node(situationA)[right = 3.0cm of partAuto, draw, rectangle, minimum width = 3.6cm, minimum height = 1.5cm, fill=gray!20] {\begin{tabular}{c}\textbf{situational}\\\textbf{awareness}\end{tabular}};
  \node(situationAI)[above left = -0.0cm and -0.0cm of situationA, anchor=north west] {V)};

  \node(urgResponse)[above = 0.3cm of situationA, draw, rectangle, minimum width = 3.6cm, minimum height = 1.5cm, fill=gray!20] {\begin{tabular}{c}\textbf{urgent response}\\\textbf{demand}\end{tabular}};
  \node(urgResponseI)[above left = -0.0cm and -0.0cm of urgResponse, anchor=north west] {VI)};
  \node(urgCommuni)[below = 0.3cm of situationA, draw, rectangle, minimum width = 3.6cm, minimum height = 1.5cm, fill=gray!20] {\begin{tabular}{c}\textbf{urgent}\\\textbf{communication}\end{tabular}};
  \node(urgCommuniI)[above left = -0.0cm and -0.0cm of urgCommuni, anchor=north west] {VII)};

  \draw [thick, draw=none](fullAuto.east)--(partAuto.west) node (rs) [circle, pos=0.5, inner sep=0pt,minimum size=1mm] {};

  \node(vStatus)[below left = 2.8cm and 0.15cm of rs, draw, rectangle, minimum width = 3.0cm, minimum height = 1.5cm] {\textbf{vehicle status}};
  \node(vStatusI)[above left = -0.0cm and -0.0cm of vStatus, anchor=north west] {I)};

  \node(nonStatus)[below right = 2.8cm and 0.15cm of rs, draw, rectangle, minimum width = 3.0cm, minimum height = 1.5cm] {\begin{tabular}{c} \textbf{non-vehicle related}\\\textbf{status} \end{tabular}};
  \node(nonStatusI)[above left = -0.0cm and -0.0cm of nonStatus, anchor=north west] {II)};
  \node(comfort)[left = 0.3cm of vStatus, draw, rectangle, minimum width = 3.0cm, minimum height = 1.5cm] {\textbf{comfort}};
  \node(comfortI)[above left = -0.0cm and -0.0cm of comfort, anchor=north west] {IV)};

  \node(nonU)[right = 0.3cm of nonStatus, draw, rectangle, minimum width = 3.0cm, minimum height = 1.5cm] {\begin{tabular}{c}\textbf{non-urgent}\\\textbf{response demand}\end{tabular}};
  \node(nonUI)[above left = -0.0cm and -0.0cm of nonU, anchor=north west] {III)};

   \node(emotion)[left = 0.4cm of comfort, draw, rectangle, minimum width = 3.0cm, minimum height = 1.5cm, dashed, fill=gray!20]{\begin{tabular}{c}\textbf{emotion and}\\\textbf{relaxation regulation}\end{tabular}};
   \node(emotionI)[above left = -0.0cm and -0.0cm of emotion, anchor=north west] {IX)};

  \node(trust)[left = 3.4cm of fullAuto, draw, rectangle, dashed, minimum width = 3.6cm, minimum height = 1.5cm, fill=gray!20] {\begin{tabular}{c}\textbf{building}\\\textbf{trust}\end{tabular}};
  \node(trustI)[above left = -0.0cm and -0.0cm of trust, anchor=north west] {VIII)};

  \draw [-{Latex[length=2mm, width=2mm]}, dashed, thick, draw=black](fullAuto.west)--(trust.east);
  \draw [ dashed, thick, draw=black](emotion.north)--(trust.south);

   \draw [-{Latex[length=2mm, width=2mm]}, dashed, thick, draw=black](comfort.west)--(emotion.east);
   \draw [-{Latex[length=2mm, width=2mm]}, thick, draw=black](partAuto.east)--(situationA.west);

   \node(rs1)[below = 1.0cm of rs, draw=none, rectangle]{};
   \draw [thick, draw=black](fullAuto.south)|-(rs1.center);
    \draw [thick, draw=black](partAuto.south)|-(rs1.center);
   \draw [-{Latex[length=2mm, width=2mm]}, thick, draw=black](rs1.center)-|(comfort.north);
   \draw [-{Latex[length=2mm, width=2mm]}, thick, draw=black](rs1.center)-|(vStatus.north);
   \draw [-{Latex[length=2mm, width=2mm]}, thick, draw=black](rs1.center)-|(nonStatus.north);
   \draw [-{Latex[length=2mm, width=2mm]}, thick, draw=black](rs1.center)-|(nonU.north);

   \draw [thick, draw=black](situationA.north)--(urgResponse.south);
   \draw [thick, draw=black](situationA.south)--(urgCommuni.north);

   \draw [decorate,decoration={brace,amplitude=5pt, aspect=0.50}, draw=black, ultra thick] let \p1=($(nonU.south east)$), \p2=($(comfort.south west)$) in ($(\x1,\y2)+(0,-0.1)$)-- node[below=0.2cm] (controlCBrace) {\textbf{interaction clusters for assistant driving} (level 0 and 1)} ($(\x2,\y2)+(0,-0.1)$);
    \draw [decorate,decoration={brace,amplitude=5pt, aspect=0.50}, draw=black, ultra thick] let \p1=($(urgCommuni.south east)$), \p2=($(comfort.south west)$) in ($(\x1,\y2)+(0,-1.0)$)-- node[below=0.2cm] (controlCBrace) {\textbf{interaction clusters for partial automated driving} (level 2 and 3)} ($(\x2,\y2)+(0,-1.0)$);
   \draw [decorate,decoration={brace,amplitude=5pt, aspect=0.50}, draw=black, ultra thick] let \p1=($(nonU.south east)$), \p2=($(emotion.south west)$) in ($(\x1,\y1)+(0,-1.9)$)-- node[below=0.2cm] (controlCBrace) {\textbf{interaction clusters needed for fully automated driving} (level 4 and 5)} ($(\x2,\y1)+(0,-1.9)$);

  \end{tikzpicture}}
\caption{Overview of interaction clusters I)--IX) for assistant, partial automated, and automated driving, i.e., level 0 to 5; gray boxes indicating important driver/passenger UI and HMI; dashed boxes are currently neglected in the development of automated vehicles.}\label{fig:overview}
\end{figure*}

Considering the needs of the passenger in fully automated vehicles and the needs of both the diver and the passenger in partially automated vehicles, nine interaction clusters can be defined.
As illustrated in Fig.~\ref{fig:overview}, seven interaction clusters, I) to VII), are essential for partial automation according to levels 2 and 3 of the SAE J3016, and six interaction clusters, I) to IV), VIII) and IX), for automated driving according to levels 4 and 5.

Within the well-known interaction clusters I) and II), cluster I) informs the driver and the passengers about the vehicle status, like the remaining driving range, and cluster II) about non-vehicle related status, like driving distance to the destination. Along with non-urgent responses II) like maintenance information and multimedia settings as part of the cluster comfort IV), these four clusters are available in current vehicles from 0 to 3 and will be present in automated vehicles, i.e., levels 4 and 5. For partial automation levels 2 and 3, the interaction clusters situational awareness V), i.e., making the driver aware of obstacles, urgent response VI) demand like the takeover from the autopilot to the driver and urgent communication VII) like error state information are vital only to the drivers. Currently mostly neglected are the interaction clusters of building trust VIII) in the system and regulating emotion and relaxation IX). However, two clusters VIII) and IX) are important since in levels 4 and 5, only passengers are in the vehicles.

Based on these nine interaction clusters, existing literature on how UI and HMI interact with users is reviewed in the remaining and summarized in Table~\ref{tab:comparisionAll}. There are numerous different modalities used for user interaction. These modalities are used for various use cases within the vehicle and rely on one single modality, cf. Section~\ref{monoM}, or multiple modalities cf. Section~\ref{muliM} to engage with users. Each modality has its strengths and weaknesses. For instance, auditory output produces better reaction times, while visual output improves drivers' efficiency. It is generally accepted that multimodal interventions \cite{Bengler2020} have a higher impact as their reaction and process time are shorter and lead to fewer error-prone actions.

\section{Mono-Modal User Interfaces} \label{monoM} 
Mono-model UI uses visual, auditory or haptic channels to communicate with the passengers and drivers---the olfactory channel has not yet been used in the context of self-driving vehicles. However, olfactory UI is already taken into account for recognizing the system status of mainframes \cite{Schoening2022}. The following sections summarize all reviewed UI and HMI from Table~\ref{tab:comparisionAll}; the corresponding citations for each section can be found there.

\subsection{Visual}
Visual UI uses text, graphics, augmented reality (AR), and lights to communicate verbal and non-verbal.
\subsubsection{Text}
Much information is typically transmitted via text during the interaction between the driver and passenger. Text UI and HMI can be employed both to alert users and provide non-urgent information. Text is processed faster than unknown graphics and icons; however, the user must be capable of reading and understanding the text. As a result, the use of alerting words like \textit{Danger}, \textit{Warning}, and \textit{Error} is helpful to users. The text allows for comprehensive and detailed communication with users. Large display areas are needed to ensure a readable character size in these cases.

\vfill
\subsubsection{Graphic and Icons}
Graphics and icons allow users to communicate information effectively in limited display areas. Graphics and icons with emotional components are processed faster than neutral images. Thus, when using these methods for urgent communication, it is suggested that emotionally arousing graphics and icons should be used in the design. Users need domain knowledge for some graphics and icons, like the brake and the headlights icons; thus, processing unknown pictures might become infeasible.

\vfill
 \subsubsection{Augmented Reality}
AR is used as a form of dynamic output to communicate with users. It is usually employed to increase situational awareness and drive performance. However, it combines real-world items with projected texts, graphics, and icons. Headup displays can be used as AR displays but are used primarily as transparent displays that do not embed real-world items in the HMI.

\vfill
\subsubsection{Light} Various implementations of light have been used in HMI. They are used in vehicles' dashboards, glasses, and general light fixtures. Lights are a dynamic output for alerting users about vehicle status, increasing situational awareness, and communicating urgent messages.

\begin{table*}[!ht]
\caption{Comparison of the HMI, categorized by the primarily triggered modality;\\ the level are based on the SAE J3016.} \label{tab:comparisionAll}
\centering
\resizebox{\linewidth}{!}{
\renewcommand{\arraystretch}{1.75}
\begin{small}
\begin{tabular}{|>{\centering\arraybackslash}p{0.085\linewidth}|
>{\raggedright\arraybackslash}p{0.25\linewidth}|
>{\raggedright\arraybackslash}p{0.45\linewidth}|
>{\centering\arraybackslash}p{0.1\linewidth}|
>{\centering\arraybackslash}p{0.15\linewidth}|
>{\centering\arraybackslash}p{0.095\linewidth}|
}
\hline
\textbf{Modality} & \textbf{Goal} & \textbf{Application}  & \textbf{SAE Level} & \textbf{Intera. Cluster} & \textbf{Reference} \\\hline\hline

\multirow{17}*{\rotatebox[origin=c]{90}{\textbf{Visual} \hspace{2cm}}}

& communicating uncertainty
& graphic representation of level of uncertainty
& 3
& V)
& \cite{Helldin2013} 
\\\cline{2-6}

& winding down after work
& ambient lights
& 3
& IV)
& \cite{Terken2013} 
\\\cline{2-6}

& level of automation 
& informative dashboard
& 3
& I)
& \cite{Gowda2014} 
\\\cline{2-6}

& communicating automation intentations
& graphic representation
& 3
& VII)
& \cite{Haeuslschmid2017} 
\\\cline{2-6}

& navigating conditional automation
& graphic and text
& 3
& V)
& \cite{Forster2016} 
\\\cline{2-6}

& increasing situational awareness
& gamified driving through augmented reality
& 3
& V)
& \cite{Schroeter2016} 
\\\cline{2-6}

& increasing trust
& LED strip at the bottom of the windowpane indicating obstacles \linebreak Attention tracking  and augmentation of obstacles using eye-tracking
& 3
& VIII)
& \cite{Lungaro2017} 
\\\cline{2-6}

& situational awareness
& glasses with peripheral lights
& 3
& V)
& \cite{Veen2017} 
\\\cline{2-6}

& increasing focus
& augmentation of traffic objects
& 3
& III)
& \cite{Wintersberger2017} 
\\\cline{2-6}

& communicating reliability
& LED strip with different colors indicating different levels of reliability
& 3
& VII)
& \cite{Faltaous2018} 
\\\cline{2-6}

& communicating uncertainty
& LED stripe
& 3
& VII)
& \cite{Kunze2018} 
\\\cline{2-6}

& improving usability
& AR HUD
& 3
& VI)
& \cite{Schoemig2018} 
\\\cline{2-6}

& drawing driver attention
& LED stripe
& 3
& V)
& \cite{Troesterer2018} 
\\\cline{2-6}

& improving TOR performance
& early 3D AR representation of traffic situation ahead
& 3
& VI)
& \cite{Wiegand2018} 
\\\cline{2-6}

& improving situation awareness
& abstract graphics representing system’s elements
& 3
& V)
& \cite{Wiegand2019} 
\\\cline{2-6}

& reliability communication
& Led strip with different colors indicating different levels of reliability
& 3
& VII)
& \cite{Loecken2020} 
\\\cline{2-6}

& improving driver’s trust
& continuous LED blue light indicating  automation, pulsing red light indicating TOR and pulsing blue light indicating availability of the automation
& 3
& III)
& \cite{Louw2021} 
\\ \hline

\end{tabular}
\end{small}
}
\end{table*}

\setcounter{table}{0}
\begin{table*}[!ht]
\captionsetup{justification=centering}
\caption{Comparison of the HMI, categorized by the primarily triggered modality;\\ the level are based on the SAE J3016. (continued)} \label{tab:comparisionAll2}
\centering
\resizebox{\linewidth}{!}{
\renewcommand{\arraystretch}{1.75}
\begin{small}
\begin{tabular}{|>{\centering\arraybackslash}p{0.085\linewidth}|
>{\raggedright\arraybackslash}p{0.25\linewidth}|
>{\raggedright\arraybackslash}p{0.45\linewidth}|
>{\centering\arraybackslash}p{0.1\linewidth}|
>{\centering\arraybackslash}p{0.15\linewidth}|
>{\centering\arraybackslash}p{0.095\linewidth}|
}
\hline
\textbf{Modality} & \textbf{Goal} & \textbf{Application}  & \textbf{SAE Level} & \textbf{Intera. Cluster} & \textbf{Reference} \\\hline\hline

\multirow{8}{*}{\rotatebox[origin=c]{90}{\textbf{Visual-Auditory}\hspace{1.5cm}}}

& winding down after work
& calming auditory and visual theme
& 3
& IX)
& \cite{Terken2013} 
\\\cline{2-6}

& improving takeover alert
& visual information, earcons
& 3
& VI)
& \cite{Walch2015} 
\\\cline{2-6}

& improving TOR
& auditory cues, ambient lights in peripheral vision(steering wheel)
& 3
& VI)
& \cite{Borojeni2016} 
\\\cline{2-6}

& handling cooperative driving better
& graphic and text alerts, auditory alerts
& 3
& V)
& \cite{Walch2016} 
\\\cline{2-6}

& increasing trust
& avatar with direct dialogues with the driver, representing internal states  with led lights,  gaze behavior and symbols  indicating object recognition
& 3
& VIII)
& \cite{Zihsler2016} 
\\\cline{2-6}

& improving trust and acceptance
& gong sound, voice message, text message
& 3
& IV)
& \cite{Muthumani2020} 
\\\cline{2-6}

&improving driver’s understanding of vehicles  when automation is activated
& auditory cues, driving centric/vehicle centricvisual panel
& 1/2
& I)
& \cite{Monsaingeon2021} 
\\\cline{2-6}

& improving attention and situational awareness
& LED lights conveying intention of the automation, threat and TOR by color and  movement and pulsing
& 3
& V)
& \cite{Yang2018} 
\\\hline

\multirow{2}{*}{\rotatebox[origin=c]{90}{\begin{tabular}{c}\textbf{Auditory-}\\ \textbf{Haptic}
\end{tabular}\hspace{-0.075cm}}}

& conveying information about TOR situation
& auditory cues with shape-shifting steering wheel\linebreak steering wheel with vibration cues
& 3
& VI)
& \cite{Borojeni2017} 
\\\cline{2-6}

& improving TOR in drivers
& directional auditory beeps\linebreak vibrotactile alerts delivered via the seat
& 3/4
& VI)
& \cite{Petermeijer2017} 
\\\hline

\multirow{5}{*}{\rotatebox[origin=c]{90}{\textbf{Visual-Auditory-Haptics}\hspace{0.9cm}}}

& takeover communicating
& haptic cueing with tactors, speech warning, graphic Text
& 3
& VII)
& \cite{Politis2015} 
\\\cline{2-6}

& increasing usability
& AD-HMI, LED feedback, haptic feedback through seat
& 3
& IV)
& \cite{Schoemig2018} 
\\\cline{2-6}

& improving efficiency and decreasing errors
& visual feedback in the form of AR\linebreak Auditory feedback cueing the acceptance of input by driver,  haptic feedback by resistance to dangerous decisions by driver and vibration
& 4
& V)
& \cite{Manawadu2017} 
\\\cline{2-6}

& improving response to unscheduled TOR
& LED light in colors of blue, orange and red indicating availability and engagement of automation, visual icons for information communication, indicating Auditory feedback cueing the acceptance of input by driver, haptic feedback by resistance to dangerous decisions by driver and vibration
& 3
& VII)
& \cite{Manawadu2018} 
\\\cline{2-6}

& improving control transition
& audio cues, haptic seat, HUD and LED stripe  used for TOR and HOR
& $ \approx $ 3
& VI)
& \cite{Salminen2019} 
\\\hline

\end{tabular}
\end{small}
}
\end{table*}
\clearpage

\subsection{Auditory}
Verbal and non-verbal UI via the ears are performed with speech, music, and earcons.

\subsubsection{Earcon} These are often used for brief and urgent user communication. Although it is not possible to use them for detailed communication, they have the advantage of incurring a low cognitive load. Like other auditory output mediums, it is possible to arouse users' attention to them when they are not actively engaged with driving tasks.
\subsubsection{Speech} Although not preferable for urgent situations due to the high cognitive load it needs, speech has the distinct benefit of anthropomorphizing interaction with the vehicle and creating a welcoming atmosphere inside the vehicle.
\subsubsection{Music}
Music is almost exclusively used as an intervention for emotional regulation.

\subsection{Haptics}
Haptics are almost exclusively used for non-verbal UI.
\subsubsection{Vibration}
Most vibrations in vehicles are delivered through vehicle seats, wearable technology, and steering wheel. Vibrations are an excellent medium of brief communication with users.
\subsubsection{Steering Wheel dynamics} Shape-shifting steering wheels and resistance of the wheel to dangerous choices by drivers are some of the ways steering wheels have been used to communicate with users.

\section{Multi-Modal User Interfaces }\label{muliM}
The previous overview of mono-modal interfaces, as well as Table~\ref{tab:comparisionAll}, makes it clear that multisensory processing is an area of interest for the interface designs of the interaction clusters VIII) and IX) in self-driving vehicles. Thus, this section briefly reviews the theoretical framework in this field. The research of a multisensory perspective toward perception goes back to the Gestalt movement. The Gestalt movement resulted from a backlash against the dominance of the reductionist theories. The Gestalt movement circles the idea that perception should be viewed as a holistic phenomenon \cite{jansson-boyd_routledge_2017}.

The idea of a holistic phenomenon is translated into one of the central notions of Gestalt theory, namely holism or the principle of totality. According to this principle, a sensory whole transcends its parts and becomes something else in its quality \cite{wagemans_century_2012}.
Gestalt theory paved the way for research on multisensory integration. Stein and Meredith \cite{steinicke_human_2013} introduced integrative rules for multisensory integration. The first rule, the spatial rule, states that spatial coincidence enhances the responses of multimodal neurons while spatial non-alignment depresses their responses. The second or temporal rule suggests that temporal proximity enhances neural responses. The final rule, or the inverse effectiveness rule, proposes that the magnitude of response enhancement is inversely related to the intensity of a stimulus. A stimulus with a weak intensity results in a more significant response enhancement than a stimulus with high intensity \cite{liversedge_oxford_2011}.

One problem with sensory processing is how noisy neural processes are. As a result, an essential component of these processes has to be the reduction of the variance to create the best sensory estimates. One influential model that strives to explain this outcome in the multisensory integration processes is the maximum likelihood estimation model based on forced-fusion and Gaussian assumptions. This model proposes that multisensory integration is based on the weighted average of the signal and that these weights are consistent with the relative reliabilities of the signals \cite{steinicke_human_2013}.

In working memory, the research revolves around whether or not the information is stored in separate or integrated representations based on modality or domain. One of the pioneering work in this field is by Atkinson and Shiffrin \cite{Atkinson1968}. They proposed that after the initial modality-based processing of information, the information is stored in an amodal form in the working memory. In contrast to this model, Baddeley and Hitch \cite{Baddeley1974}, proposed a model that stores information based on its modality--the Visual-spatial sketchpad and phonological loop.

As the evidence mounted in favor of interaction between processes linked to different modalities, this model was amended, and an episodic buffer was added to the model to link the other components \cite{quak_multisensory_2015}.

Another area of interest is attention. Crossmodal sensory stimuli compete for attention. This competence could lead to the extinction of the stimuli. This phenomenon has been famously observed in the Colavita effect. In this task, participants are asked to respond to auditory and visual stimuli with different keys. However, sometimes, both stimuli are presented at the same time. Simultaneous processing of stimuli could result in failure, in which one only responds to auditory stimuli but not visual ones. Other studies have shown visual dominance in haptic vs. visual stimuli Colavita task and absence of any modality dominance in a tri-sensory task of visual, auditory, and haptic stimulus \cite{nobre_oxford_2014}.

\begin{figure}
  \centering
    \subfloat[\footnotesize emergency situation---deer ahead of the driverless vehicle \label{fig:teaser1}]{{\includegraphics[width=0.6\linewidth]{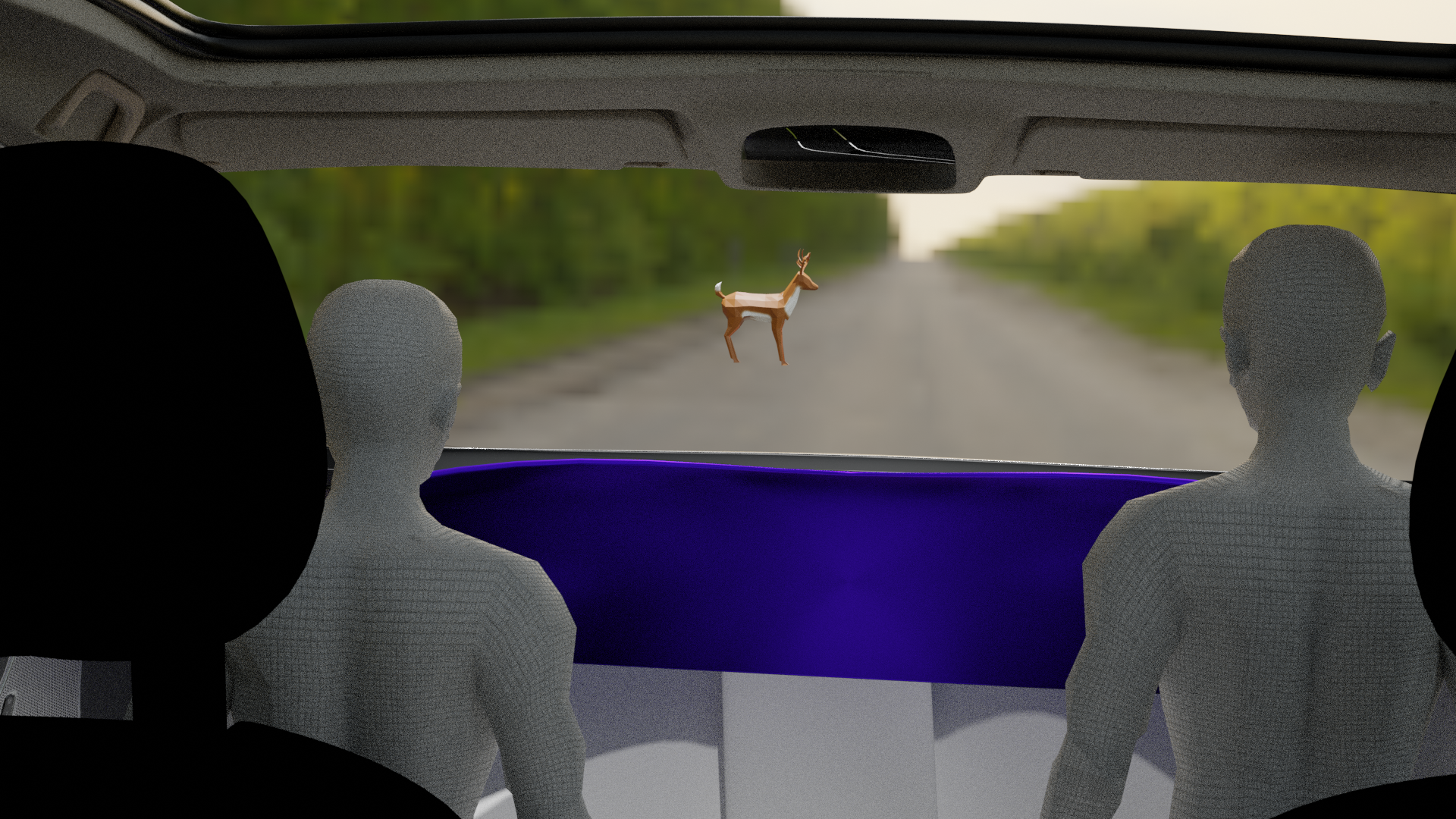} }}%
    \vspace{0.25cm}

    \subfloat[\footnotesize ambient light HMI---signaling passengers that the situation is recognized \label{fig:teaser2} ]{{\includegraphics[width=0.6\linewidth]{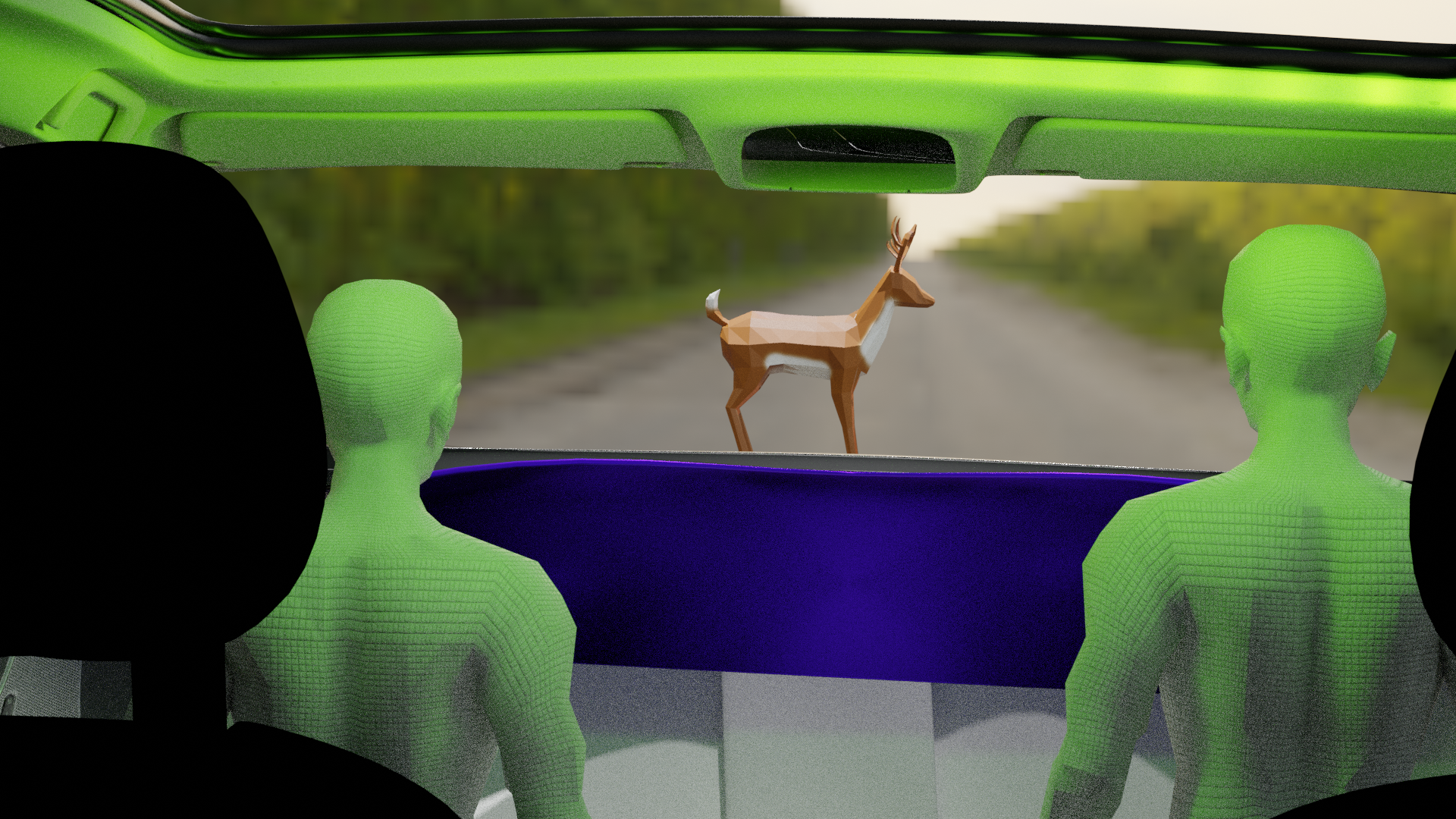} }}%
    \vspace{0.25cm}

    \subfloat[\footnotesize tactile HMI---preparing passengers for evasive manover \label{fig:teaser3} ]{{\includegraphics[width=0.6\linewidth]{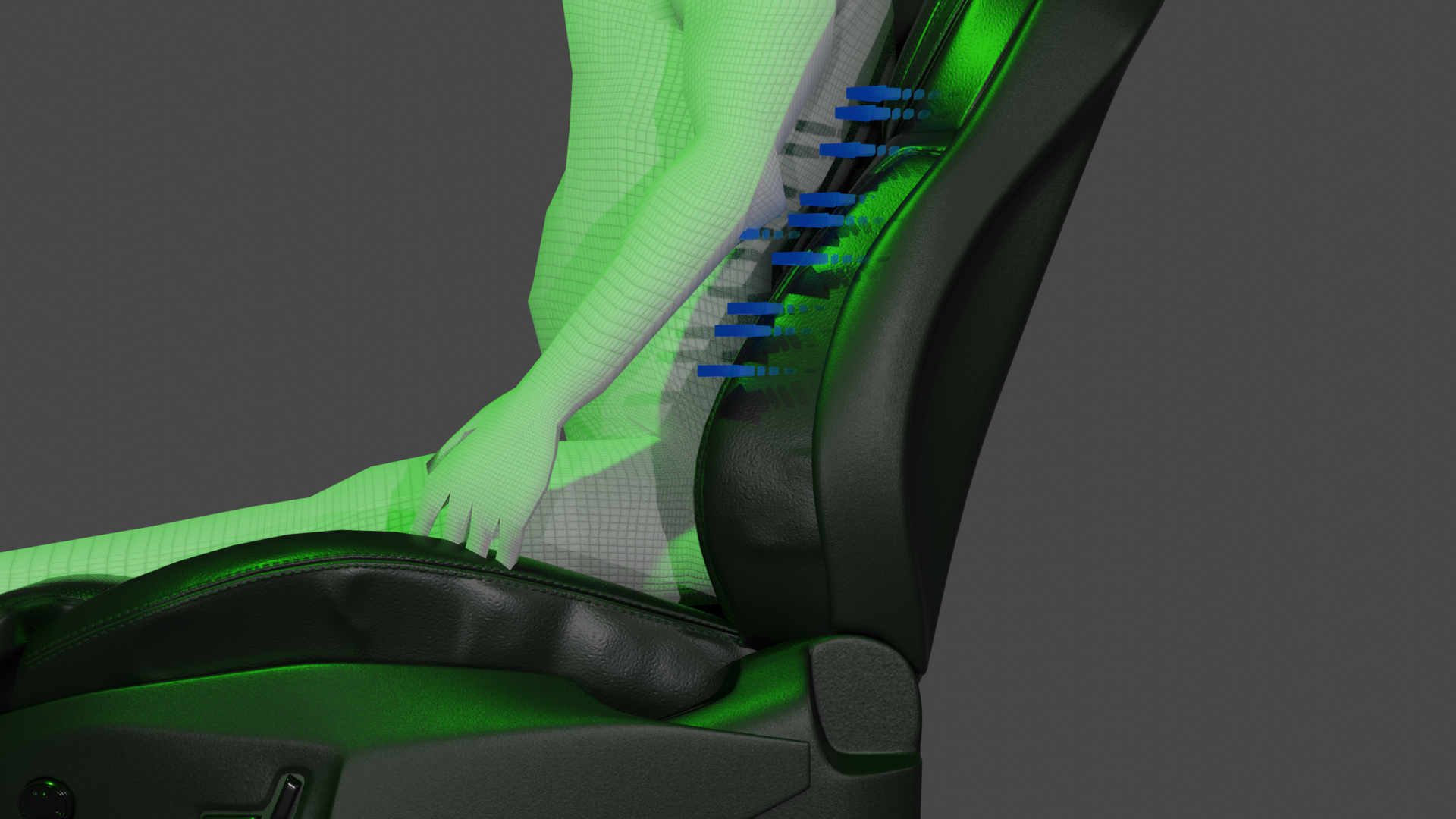} }}%
  \caption{Revolutionizing trust and comfort in self-driving vehicles, enhancing HMI for smarter mobility to tell the passengers, even in emergency situations, that everything is in control; cf. \protect\subref{fig:teaser1}---\protect\subref{fig:teaser3}.}
  \label{fig:teaser}
\end{figure}

\section{Trust and Emotion User Interfaces}\label{sec:trust}
Fully automated self-driving vehicles present unique difficulties when it comes to users' comfort, cf. interaction cluster IV), users' trust cf. cluster VIII) and users' emotions cf. cluster IX). Although the burden of driving tasks is lifted from users, it does not guarantee a comfortable experience for passengers. Focussing on unexpected events and automation surprise, i.e., the consequence of the system doing something the user did not expect \cite{Dekker2017} UI and HMI for level 4 and 5 vehicles are explored in this section.

Unexpected events and automation surprises can cause negative emotions such as stress in users \cite{Stephenson2020, Dillen2020}. Thus, examining how users handle these circumstances and regulate their negative emotions is essential. Emotion regulation refers to efforts to control what, when, and how emotions are experienced and expressed. Gross~\cite{Gross2015} divides strategies of emotion regulation into two categories of: antecedent-focused and response-focused strategies. Antecedent-focused strategies include strategies that focus on regulating emotion before the response to it emerges. These include situation selection, situation modification, attention deployment, and cognitive change. Response-focused strategies are those strategies that commence after the emotional response has already started. These strategies aim to affect emotional response's behavioral, physiological, and experiential aspects \cite{Gross1998}. These strategies could also be categorized along the dimensions of emotion regulation goal--explicit and implicit--and emotion regulation process (controlled and automatic) \cite{Braunstein2017}. Users may use any of these strategies to deal with discomfort. However, some of these strategies are maladaptive and can lead to avoidance and distrust of automation.

Moreover, interventions to help users successfully regulate their emotions could be implemented to prevent such outcomes. As reviewed by Braun et al.~\cite{Braun2019} there is a range of studies investigating possible interventions to regulate stress and discomfort during driving. These interventions use various techniques such as implicit influencing using ambient lights\cite{Hassib2019}, temperature \cite{Schmidt2019}, and music \cite{Jeon2012}. It is useful to review these strategies even if this work focuses on fully automated vehicles and improving user experience and comfort.

Higher levels of automation can present particular challenges regarding users' acceptance, trust, and comfort. It has been shown that higher automation is correlated with lower perceived control and fun \cite{Roedel2014} and lower perceived levels of safety and higher anxiety \cite{Hewitt2019}. Moreover, higher automation can jeopardize the sense of agency of users \cite{Wen2019}. The increase in automation is linked to a decrease in the sense of agency as shown by intentional bidding measures \cite{Berberian2012}. Even with interventions to improve the sense of agency, an increase in automation beyond $90\%$ strains the sense of agency \cite{Ueda2021}.

Not only a lower sense of agency and control can cause and aggravate negative emotional experiences \cite{Henderson2012}, but it can also negatively influence emotion regulation itself \cite{Delgado2008,Paolo2017}. Implementing emotion regulation strategies in self-driving vehicles starts with understanding situations that can give rise to negative emotions. One study has shown that sources of tension in highly automated vehicles can be divided into crosswalks, roundabouts, straight intersections, left-turns with on-coming traffic, left-turns without on-coming traffic, and right turns with right turns with right-of-way and right turns without right-of-way \cite{CohenLazry2020}. While this highlights the possible scenarios that can lead to discomfort, it fails to consider the factor of expected and unexpected automation action and surprise. This factor is even more significant in highly automated vehicles as It is envisioned that in highly automated vehicles, users do not attend to the road and instead engage in non-driving related tasks. As a result, it is necessary to address specific problems related to negative emotion and emotion regulation in higher automated vehicles and develop appropriate solutions for them.

Being well-informed is a possible route to reducing discomfort by heightening a sense of control during travel. For example, one study showed that a higher level of visual information inspires more trust in users of automated vehicles \cite{Ma2021}. However, it should be noted that the need for information is dynamic and highly dependent on the individual. Experience can change the level of information that users seek \cite{Ekman2018}. Overall, all interventions should be tailored and modified according to individual needs.

\section{Concepts on User Interfaces for Building Trust and Regulating Emotions}
Building trust and regulating emotions must be tailored to each passenger's needs as discussed in Section~\ref{sec:trust}. For handling the UI in emergency situations, like a sudden appearance of a deer ahead of an automated vehicle cf. Fig.~\ref{fig:teaser} (a), this section first conceptualized an ideal HMI based on the literature reviewed in Table~\ref{tab:comparisionAll} and the previous sections. Based on this concept, the research directions that are needed are discussed in the remainder.

Focusing on regulating negative emotions triggered by unexpected events and emergency situations on every single passenger, UI must focus on every single passenger's needs. Consequently, the first step is detecting signals of stress in each user. Measures of physiological and physical responses related to stress include a wide range of signals such as heart rate, skin conductance, pupil size, and respiratory rate. Measuring multiple signals is preferable as multimodal biosignal analysis allows for capturing a more accurate picture of the emotional state \cite{Lee2020}.

\begin{figure}[htp]
    \centering
    \includegraphics[width=0.5\textwidth]{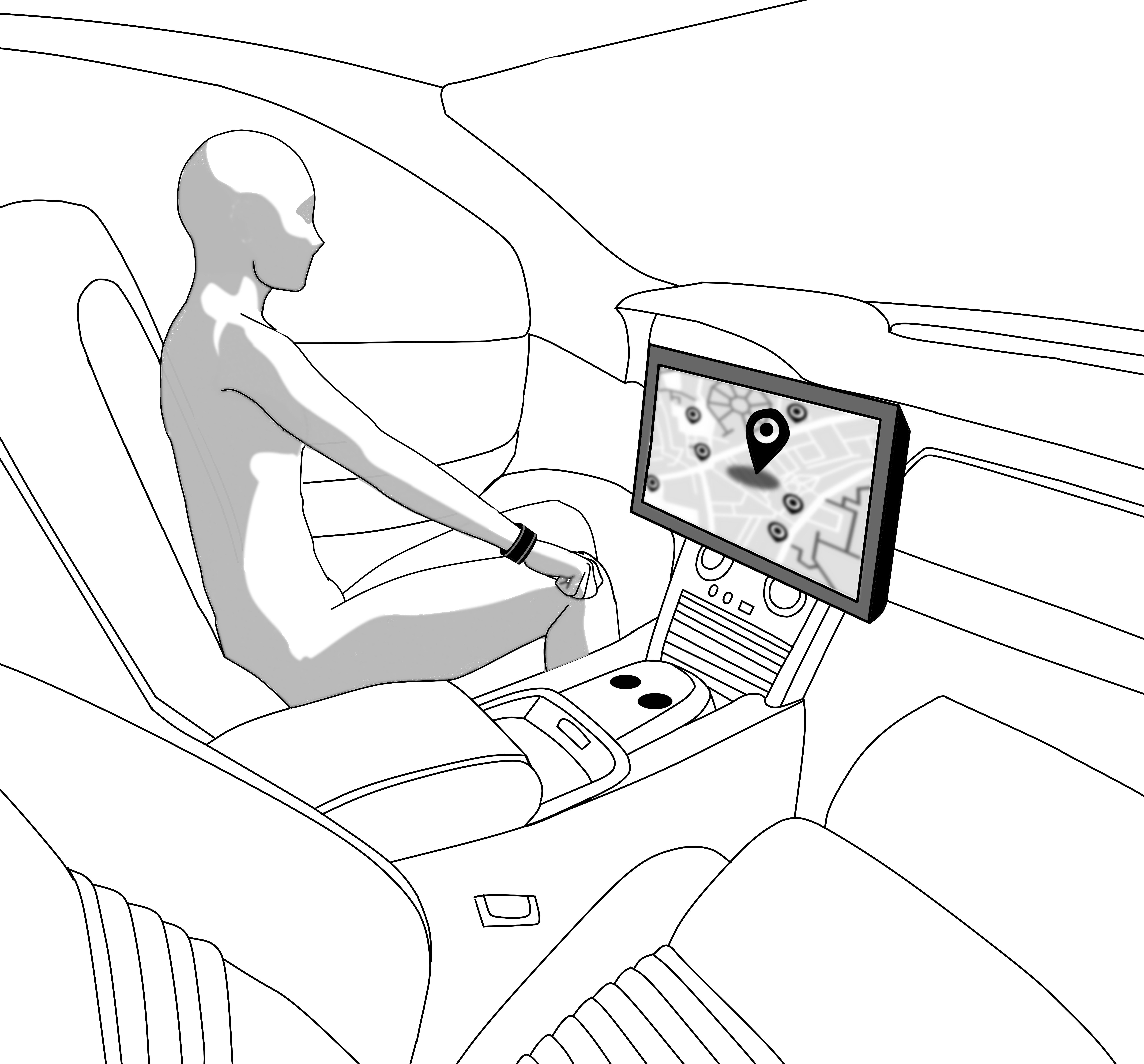}
    \caption{Building trust and regulating emotions in each passenger--with the use of a smartwatch, the individual stress levels of the passengers are monitored, and based on their emotional states, music via directional loudspeakers is played back only for each passenger. In addition, smart e-textiles softly vibrate and regulate the passengers' heat rate.}\label{fig:concept}
\end{figure}

HMI should be able to incorporate such measures and thus provide multimodal emotional regulation interventions at appropriate times for each passenger. One possible way to incorporate these measures and deliver emotional regulation intervention is by utilizing wearable technology. Wearable technology has been repeatedly used in previous research in this field. They can measure various physiological responses and are easy to use for each passenger. E-textiles are similarly capable of delivering interventions and measuring stress-related signals. As seen in Fig.~\ref{fig:concept}, a smartwatch measures the individual stress level of the passenger. Based on their emotional states first, music via directional loudspeakers is played back. Due to directional loudspeakers, each passenger can hear their anti-stress music at their preferred volume level. As the next invention, the HMI will use vibration motors in the seat, cf. Fig.~\ref{fig:teaser} (c) or in smart e-textiles. If these passenger-based emotion regulations do not show an effect, the ambient light will empower the individual UI.

With the emergence of driverless, automated, and self-driving vehicles, the importance for interaction clusters: VII) building trust and IX) emotion and relaxation regulation of Fig.~\ref{fig:overview} rises. Thus, the new UI and HMI inside these vehicles must focus on each passenger individually. The focus on each passenger is a new research direction in UI and HMI that is needed for a calm and relaxed ride within the vehicle of the future. 

\bibliographystyle{ACM-Reference-Format}
\bibliography{references.bib}


\includegraphics[]{pngSeed}

\end{document}